\documentclass[conference]{IEEEtran}

\usepackage{times}
\usepackage{epsfig}
\usepackage{graphicx}
\usepackage{amsmath}
\usepackage{amssymb}
\usepackage{times}
\usepackage{epsfig}
\usepackage{graphicx}
\usepackage{amssymb}
\usepackage{subfigure}
\usepackage{placeins}
\usepackage{afterpage}

\usepackage[breaklinks=true,bookmarks=false]{hyperref}

\renewcommand\baselinestretch{1.01}
\begin{document}
	
\title{Contextual road lane and symbol generation for autonomous driving}

\author{\IEEEauthorblockN{Ajay Soni}
	\IEEEauthorblockA{\textit{ZF TCI}\\
		Hyderabad, India \\
		ajay.soni@zf.com}
	\and
	\IEEEauthorblockN{Pratik Padamwar}
		\IEEEauthorblockA{\textit{ZF TCI}\\
		Hyderabad, India \\
		pratik.padamwar@zf.com}
	\and
	\IEEEauthorblockN{Krishna Reddy Konda}
	\IEEEauthorblockA{\textit{ZF TCI}\\
	Hyderabad, India \\
	krishna.konda@zf.com}
	}	
\maketitle

	\begin{abstract}
		In this paper we present a novel approach for lane detection and segmentation using generative models. Traditionally discriminative models have been employed to classify pixels semantically on a road. We model the probability distribution of lanes and road symbols by training a generative adversarial network. Based on the learned probability distribution, context aware lanes and road signs are generated for a given image which are further quantized for nearest class label. Proposed method has been tested on BDD100K and Baidu's ApolloScape datasets and performs better than state of the art and exhibits robustness to adverse conditions by generating lanes in faded out and occluded scenarios.
	\end{abstract}
	
	\section{Introduction}
	
	Lane detection and classification is an integral part of autonomous driving (AD) and  driver assistance (ADAS) system. In order to safely maneuver the vehicle on road and comply with driving regulations, accurate detection and segmentation of road symbols and lane is of paramount importance. Camera has been the most widely used sensor for lane detection. Early research in this area relied on edge or contour detection and consequent post processing techniques to separate out lanes and other irregular shapes on the road. In recent times traditional vision based techniques have been replaced by methods which use convolutional neural networks(CNN) for image segmentation. Usage of CNN has lead to the marked improvement in accuracy. However such approach is purely discriminative in nature and relies on pixel intensity variations and location semantics for lane segmentation. Which in turn leads to susceptibility with respect to illumination, occlusions etc. In this paper we propose a semi-generative approach to the problem where in we try to model the data distribution of road lanes and symbols. Modelled data distribution is then used as a basis for generation of lanes with given image as a condition. Generating the lanes instead of segmenting the pixels will help address the issues of illumination, occlusions, road wear and tear etc. In this context we propose a novel conditional generative adversarial network(CGAN) based model to generate road lanes and symbols
	
	\section{Related Work}
	Lane detection problem has always been a keen interest for academicians and researchers and they have proposed many methods based on classical hand crafted feature and deep learning approach. Traditional lane detection methods relied on specialized hand crafted features like \cite{jung2013efficient,tan2014novel}. Popular choices for the hand-crafted features were color-based features \cite{chiu2005lane}, bar filters \cite{teng2010real}, ridge features \cite{lopez2010robust} etc. These features are combined with a Hough transform \cite{zhou2010novel,liu2010combining} or techniques like line and spline fitting \cite{borkar2011novel} to model line segments. Authors in \cite{saudi2008fast} use hough transform as post processing technique owing to its lower complexity and hence achieve real time performance. However hough transform is only useful in case of straight lines, hence spline fitting and hyperbola fitting had been proposed \cite{aly2008real,khalifa2009vision}. In order to exploit temporal redundancy, tracking is also been used to improve the accuracy \cite{zuwhan@2008, danescu2009probabilistic}. Increase in computational power of GPU has fueled use of neural networks for almost every vision related application especially for autonomous driving . Convolutional neural networks have shown good results towards lane and symbol detection applications. In \cite{kim2014robust} authors use convolutional neural network as an alternate detection method used in case of difficult detection scenarios along with RANSAC algorithm. Another early use case of neural networks is proposed in \cite{li2016deep}
	where in convolutional neural network is used for detection of lanes and recurrent neural network is used to better localize the lanes. In \cite{lee2017vpgnet} authors propose a network which acts as a multi utility network to detects lanes, symbols, objects and also predicts vanishing point. However with respect to lanes it can only detect ego lanes which has limited usability  for lane change and other manoeuvres. Wang et al. \cite{wang2018lanenet} proposed an interesting neural network based approach where in initially a lane proposal network is used to detect lanes and later lane localization network detects lane based on lane proposals. Input to localization network is a set of point coordinates and lanes are localized irrespective of order of point coordinates. Chougule et al. \cite{choug2018ivs} used encoder-decoder based architecture for generating segmentation mask whereas identifying ego and side lane markings is done geometrically. However distinguishing individual lane instances using spatial cues is a difficult task so Neven et al. \cite{neven2018towards} used instance segmentation as means to achieve lane detection thereby directly detecting lane instances as ego or side lanes. In \cite{saudi2008fast} saudi et al propose a decoder which is branched out into classification and detection networks separately which are later merged to achieve instance segmentation. Ghafoorian et al \cite{ghafoorian2018gan} have presented a new approach where adversarial training has been used to bolster the detection accuracy of a semantic segmentation based approach.

	\section{Contribution}
\begin{figure*}[htbp]
		\begin{center}
			\includegraphics[width=0.9\linewidth]{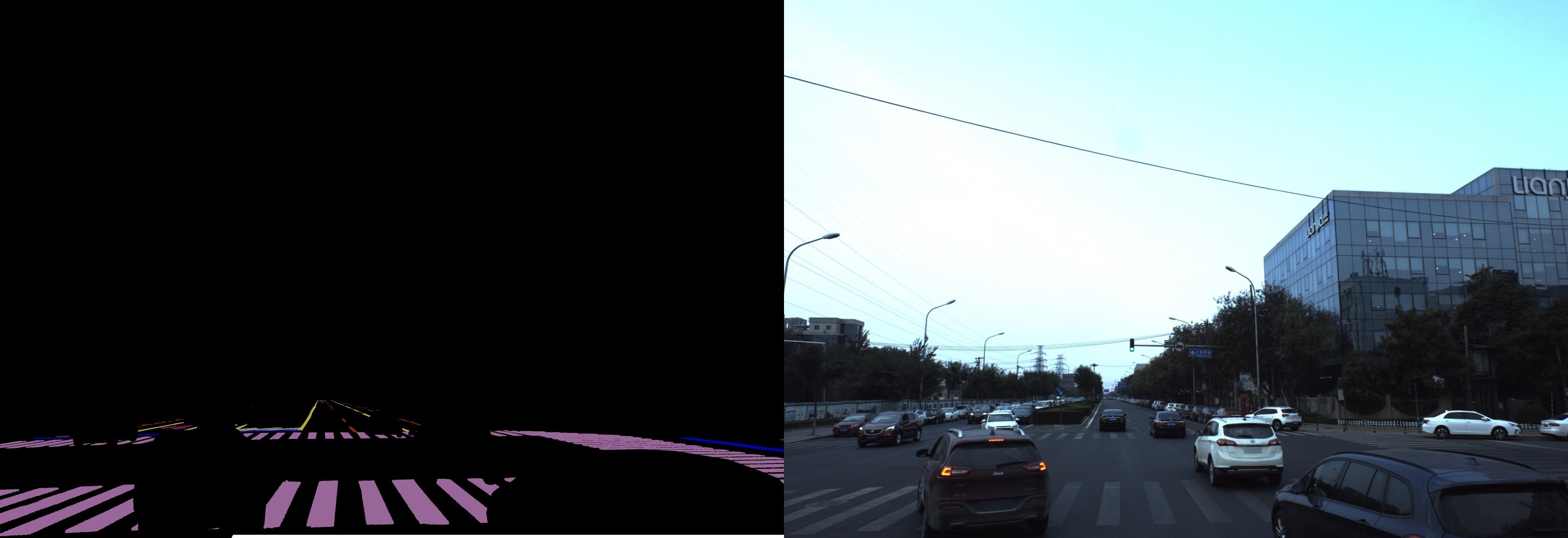}
			\caption{Faded lanes and symbols in an Image. Left-Groundtruth, Right-Camera Image}
			\label{fig:sample_faded_image}
		\end{center}
	\end{figure*}

	Despite high accuracy achieved,  current state of research mostly relies on semantics of the scene and pixel based classification. Such an approach is highly variable in performance with respect to illumination changes, occlusions and noise. In Figure \ref{fig:sample_faded_image} we have presented a case where the zebra cross markings are faded out to certain extent. In such scenarios it is very difficult for classification based networks to separate out lane pixels with that of background. However generative networks would be able to generate the lanes given that implicit data distribution has already been modeled. We demonstrate this qualitatively and also quantitatively in our results.
	
	 Disriminative models are also highly susceptible to adversarial attacks \cite{hou2020inter}. In order to address these short comings of current state of the art,  we propose to use generative modeling approach for lane detection and segmentation. By using generative networks we propose to model the probability distribution of road lanes and symbols rather than just learning discriminative features.

	Based on the modelled distribution we generate the lanes and road symbols for a given context with camera image as a conditional input to the network. Generation of lanes instead of pixel based segmentation/classification will help in detection/classification of the lanes and symbols, when there is occlusion, illumination glare etc. Given the low complexity distribution of the road lanes and symbols, proposed algorithm which relies on generative approach has a very low computational complexity when compared with state of the art segmentation based networks. The major contributions of paper can be summarized as follows:
	
	\begin{itemize}
		
		\item A generative approach which relies on modeling the lanes and road symbols rather than pixel based classification/segmentation.
		
		\item In addition to lanes proposed algorithm also generates road symbols 
		
		\item Robustness with respect to wear and tear degradation, occlusions, illumination and other noises

	\end{itemize}
	
	\section{Proposed Method}
	
	In this section we explain our approach towards generating lanes and symbol using generative models. First we explain generative neural networks  and construct our approach towards contextual lane and symbol generation using a conditioned input and conditioned discriminator.
	
	\subsection{Generative Neural Networks}
	
	Generative neural networks are class of networks which model the distribution of the given data. By mapping a given data to a known or unknown data distribution, network aims to reproduce the data for any given sample of the distribution. Prominent among the generative network models proposed are Variational auto encoders (VAE)\cite{doersch2016tutorial} and Generative adversarial networks(GAN) \cite{goodfellow2014generative}. In case of VAE any given data distribution is mapped to latent vector using an encoder which is convolutional neural network(CNN). Latent vector represents a Gaussian probability distribution. A decoding CNN is used to reproduce back the data from the modeled distribution. In case of GAN probability density function is implicit and  adversarial CNN pair is used to model the distribution using two step training where in generator generates the data and its conformance with ground truth is evaluated by discriminator CNN. GANs are being increasingly used to model the data owing to their flexibility and capability to model implicit distributions.\cite{zhu2017unpaired,isola2017image}.The objective of a GAN is illustrated in equation \eqref{GAN}

	\begin{equation}
		\begin{aligned}
			\displaystyle L_{GAN}(G,D) = \mathop{\mathbb{E}_{x{\sim}p_{data}(x)}}[log(D(x)] \\
			+ \mathop{\mathbb{E}_{z{\sim}p_z(z)}}[log(1-D(G(z)))]\
			\label{GAN}
		\end{aligned}
	\end{equation}

	Where G is the generator model and D is the discriminator. G tries to learn a data mapping from a noise prior $p_z(z)$ to data space $G(z,: \theta)$. Discriminator's job is to check weather the generated sample belongs to x rather than $p_g$ and outputs a single probability value.
	
	\subsection{Conditional GAN}
	
	A GAN can learn any implicit probabilistic distribution using the adversarial loss. But learning the distribution only helps generating random samples. However in real world applications we would need to generate samples from a class in a given context rather than generating randomly. CGAN \cite{mirza2014conditional} solved this issue by applying a condition y to the noise at the generator and discriminator input. At inference time the generator creates samples based on input condition. The objective of a CGAN thus is illustrated in equation \eqref{CGAN}
	
	\begin{equation}
		\begin{split}
			\displaystyle L_{cGAN}(G,D) = \mathop{\mathbb{E}_{x{\sim}p_{data}(x)}}[log(D(x|y)] \\
			+ \mathop{\mathbb{E}_{z{\sim}p_z(z)}}[log(1-D(G(z|y)))]\
			\label{CGAN}
		\end{split}
	\end{equation}

	\subsection{Contextual Lane and Symbol Generation}
	
\begin{figure*}[htbp]
		\begin{center}
			\includegraphics[width=0.6\linewidth]{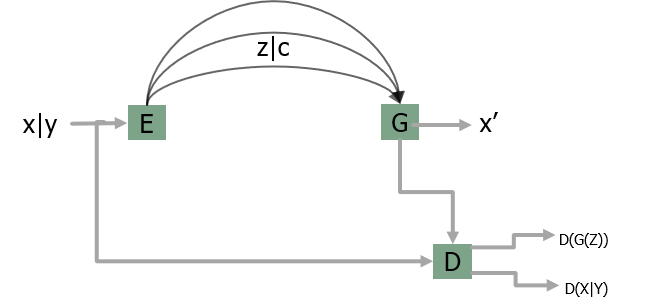}
			\caption{{Block diagram of the network}}
			\label{fig: block diagram}
		\end{center}
	\end{figure*}

	Conditional GAN (CGAN) \cite{mirza2014conditional} can learn and generate samples conditionally from a learned distribution. However CGAN's learning objective makes it learn the attributes globally and fails to capture local semantics and fine contextual information. Our work further explores the basic idea of modelling the probabilities given a context but without losing the semantics of the scene. We found that one way to achieve this is to omit the conditioning of the latent dimensions at the generator input, instead we give the contextual information(road scene image) to the encoder. In this way the generator gets conditioned information and requires no further conditioning. Also we use mean square error loss for comparing generator's output so that generator not only learns data distribution (lanes and symbols in our case) but also generates samples as close to ground truth as possible. In Figure \ref{fig: block diagram} we illustrate our concept to model the learning objective. In \eqref{conextGan} we illustrate our objective function.
	
	\begin{equation}
		\begin{split}
			\displaystyle L_{CGAN}(G,D) = \mathop{\mathbb{E}_{x,y}}[log(D(x|y)] \\
			+ \mathop{\mathbb{E}_{x,z}}[log(1-D(x|(G(x|z))))]\
			\label{conextGan}
		\end{split}
	\end{equation}
	
	\subsection{Problem formulation}
	\label{prob}
	For any given geographical location road lanes and symbols are very well defined and form a data distribution which can be closely modeled by variational autoencoders or GAN.

	Let the data distribution be represented by $\pi$ which is a function of various factors such as geographical location, type lanes etc. Let the image captured by the camera at any time instant  $I$. A generative model $G$ should generate lanes and symbols output $O$ with $I$ as condition. Hence the output $O$ can be modelled as a conditional distribution given by \eqref{generator}. where $\hat{\pi}$ is the implicit distribution represented by the proposed generator model.

	\begin{equation}
		\displaystyle O= G(\hat{\pi},I)
		\label{generator}
	\end{equation}

	For the generator to generate meaningful samples or in other words to learn an implicit probability distribution an adversary can be used. We employ a discriminative model which uses the condition $I$ as well to get an additional information about the context. This discriminator acts as an adversary to the generative model. Output $D_i$ of an adversarial network can be expressed as in equation \eqref{disc}. 
	
	\begin{equation}
		\displaystyle D= D[\pi|(G(\hat{\pi},I),I)]
		\label{disc}
	\end{equation}

	In the current context both generator network $G(\hat{\pi},I)$ and discriminator network $D$ are modeled as convolutional neural networks. 
	
	\subsection{Losses}
	
	\subsubsection{Generative Loss}
	
	The generator needs to generate image pixels based on the learned distribution. The generated image needs to be as close to ground truth as possible. To achieve this we use mean square error loss. As we are also using original ground truth as an input $\pi $ is encoded with context $I$ The generative loss takes care of quantitative aspect of the objective function.
	
	\begin{equation}
		\displaystyle L_{mse}(\theta_g) = \frac{1}{N}\sum_{i=1}^{N} || G_{\theta_g}(\hat{\pi}_i, I_i) - \pi_i||^2 
		\label{genloss_Eq}
	\end{equation}
	
	where $\pi$ is data distribution of lanes and symbols,  $i$ is the pixel location and $N$ represents the number of pixels in an image.  $I_i$ is an pixel context at location i. $G_{\theta_g}(\hat{\pi}_i, I_i)$ represents generated image pixel at location i and $\theta_g$ are its parameters.

	\subsubsection{Adversarial loss}
	The adversarial loss takes care of qualitative aspect of objective function.The generator's job is to fool the discriminator by generating random samples or pixel values. Adversarial network identifies whether the generated image looks real or not. Further as explained previously the discriminator is conditioned on road scene image hence also checks the generated image association to real road scene.
	
	\begin{equation}
		\displaystyle L_{adv}(\theta_d, \theta_g) = \frac{1}{N}\sum_{i=1}^{N} ||1-D_{\theta_d}(G_{\theta_g}(\hat{\pi}_i, I_i))||^2
		\label{advloss}
	\end{equation}
	
	Here $D_{\theta_d}(G_{\theta_g}(\hat{\pi}_i, I_i))$ represents the probability of generated pixel belonging to $\pi$. $\theta_g$ and $\theta_d$ are generator and discriminator parameters respectively. Instead of using a binary value of 0 and 1 for adversarial loss we have used rather soft values specifying degree of $G_{\theta_g}(\hat{\pi}_i, I_i)$ being drawn from a distribution of $\pi$.
	Both the networks are trained alternatively in steps to model the distribution based on losses discussed. 
	
	\section{Network architecture}
	
	We detail out the basic building blocks of our network architecture in the subsection below.
	
	\subsection{Input Conditioning and Generator}
	
	\subsubsection{Input conditioning}
	
\begin{figure*}[htbp]
		\begin{center}
			\includegraphics[width=0.93\linewidth]{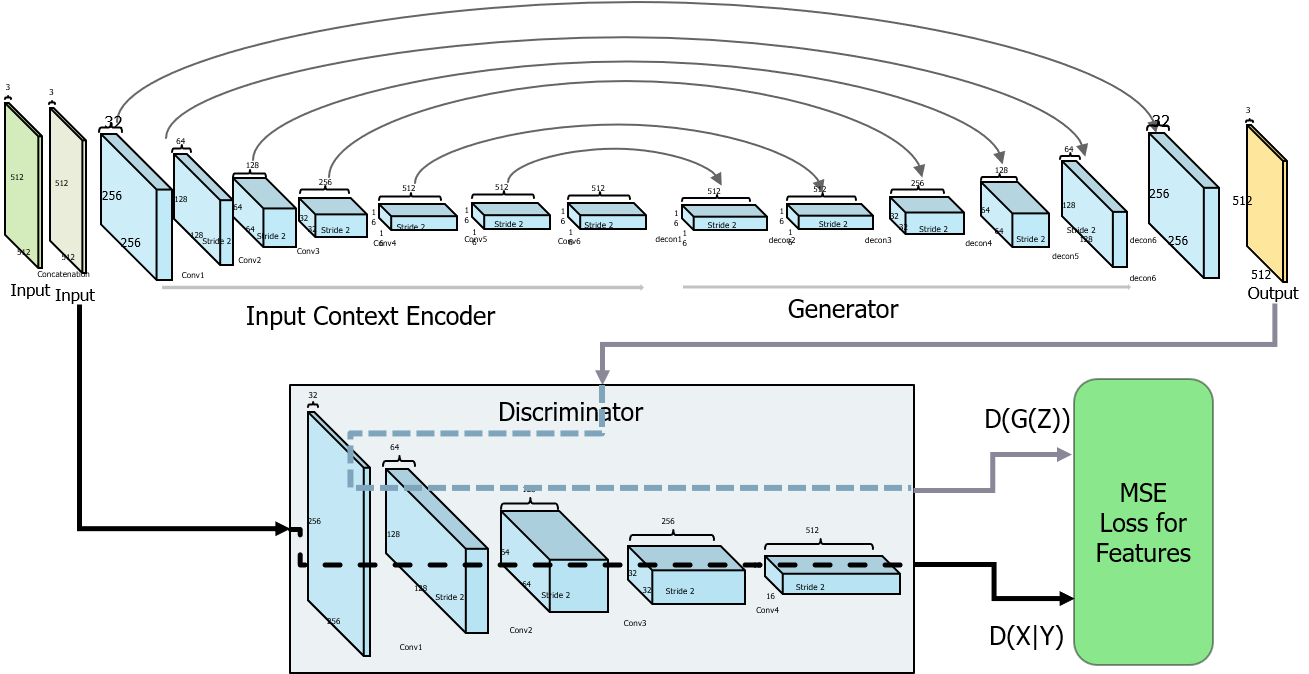}
			\caption{{Network architecture}}
			\label{fig:architecture}
		\end{center}
	\end{figure*}

	As described in the subsection \ref{prob} our aim is to conditionally generate lanes and symbols in a given image context. Subsequently input image is transformed into latent dimension to condition the generator using a classic encoder structure. Encoder transforms the input image into a feature based latent dimension which is then given as an input to generator. Similarly latent dimension vector generated from input image is also used as condition at discriminator end. However single latent dimension based conditioning leads to the loss of fine grained features present in the input image. To overcome loss of fine features conditioning of generator is done at multiple latent dimensions using skip connections. Addition of conditioning at various feature map resolutions has been arrived through extensive experimentation with respect to network architecture. Proposed architecture is shown in the Figure \ref{fig:architecture}  .  Basic encoder structure consists of Convolution $\,\to\,$ Batch Normalization $\,\to\,$ Leaky ReLU layers. Generator in turn is derived from a decoder structure which consists of Up sampling $\,\to\,$ Convolution $\,\to\,$ Batch Normalization $\,\to\,$ LeakyReLU layers. 
	
	The input to encoder is target image concatenated with color road scene image. The target as well as road scene image sizes are $512\times512\times3$. the concatenation happens along the channels hence input size is $512\times512\times6$. For learning a distribution only a decoder is sufficient as in \cite{goodfellow2014generative}. At inference time contextual encoder is given only contextual information and generator outputs lanes and symbol image based on road context.
	
	\subsection{Discriminator}
	
	In a  classic conditional GAN job of the discriminator is to asses the images generated by the generator for their similarity with respect to expect output and subsequently binary cross entropy loss is used\ref{GAN}. However in the present scenario we need both qualitative and quantitative assessment of lanes and symbols generated with respect to the ground truth. In order achieve such an objective we have replaced the binary cross entropy loss with normalized feature level mean square error loss(MSE as shown in the Figure \ref{fig:architecture}.  For discriminator we use similar architecture as for encoder. At Input we resize the input and contextual image to $512\times512\times3$ and concatenate them. The input to discriminator thus is target label conditioned on color image of road scene and the input size is $512\times512\times6$.

	\section{ Class distribution model}
	
	By definition generative model relies on data modeling the data distribution. Hence the proposed generator is expected to generate probability distribution of classification labels for each of the classes based on the context set by input image. Traditional one hot encoding for classification of labels used in case of  discriminative models cannot be used in this context owing to the discontinous nature of such a distribution. Mapping a discontinous distribution to the lane and symbol distribution which is gaussian in nature is a diffcult preposition.

	Hence we propose to map an RGB image as an output distribution owing to its inherent gaussian nature.  Which is later mapped to classification labels through postprocessing step. Where in generated pixels are quantized to nearest pre defined ground truth index image levels. One of the limitation of this approach is that the generated labels are sampled from an implicit probability hence there is no guarantee that they will belong to one of the pre-defined indices. But we see this as an advantage over discriminatory models as limiting CNN to a particular class composition stops it to exploit the inherent distribution and always tries to fit in a certain class because of which generalization is difficult to achieve.		
	
	\section{Experimental set up}
	We have tested our algorithm on BDD100K \cite{fisherYU2018BDD100K} and Baidus’s ApolloScape \cite{wang2019apolloscape} lane segmentation datasets. One of the primary reasons for choosing these dataset is that they incorporate per pixel lane and symbols semantic labels. As we are modeling the implicit probability distribution of a particular class both the datasets suit our requirements.
	
	BDD100K is a challenging dataset because of variability in road scene, illumination and lane categories. The dataset is of total 100K images with 70K for training, 10K for validation and 20K for testing. As the ground truth labels for test set are unavailable we tested our algorithm on validation set and used 10K out of 70K of train set for validation. Image resolution is $1280\times720$ but we resized the image to  $512\times512$ speed up the experiments. 
	
	We also used  Baidu’s ApolloScape lane segmentation dataset \cite{wang2019apolloscape} for our experiments. The dataset is quite large, diverse and has all the major lane and symbol classes required for city driving.
	
	The dataset contains images captured by  two front facing cameras “Camera 5” and “Camera 6”. While both the cameras are front facing there are some differences in the road area covered by each of them hence the images also have subtle differences. To avoid loss of generality we have used both “Camera 5” and “Camera 6” images for training. We have created a subset of ApolloScape dataset which comprise 10000 training images, 10000 validation and 3000 testing images that hold all major classes of road symbols and road lanes as per \cite{wang2019apolloscape}  
	
	We train the network using stepwise gradient first for D and then for G. To optimize our objective we use mini batch of 16 and use ADAM optimizer with learning rate =  0.0002 and  $\beta_1 = 0.5$. We trained our model for various epochs and see generator loss stabilized around 150 epochs.
	
	For inference, we create an image with Gaussian noise of size $512\times512\times3$ and concatenate it with real world image of same size for contextual information and run generator in the same manner as in training.
	
	\begin{table}[htbp]
		\small
		\centering
		\caption{Dataset summary for experiments}
		\begin{tabular}{|c|c|c|}
			\hline
			$Attribute$& $BDD100K$ & $Apolloscape$ \\
			\hline\hline
			Frames & 80000 & 15000  \\
			Train & 60000 & 10000 \\
			Val  & 10000 & 2000  \\	
			Test  & 10000 & 3000 \\	
			Original Size   & 3384 x 1710 & 1280 x 720  \\	
			Training Size  & 512 x 512 & 512 x 512 	\\
			Road Scene  & Urban, rural, highway & Urban \\	
			Classes & 8 & 10   \\		
			\hline
		\end{tabular}%
		\label{tab:experiments}%
	\end{table}%
	
	\section{Results}
		\begin{table}[htbp]
		\small
		\centering
		\caption{State of the art comparison for Apolloscape dataset}
		\begin{tabular}{|c|c|c|}
			\hline
			S no & Method & mIOU \\
			\hline\hline
			1 & Baseline Wide ResNet-38   & 0.422  \\ 
			2 & UNet-ResNet-34   & 0.424 \\ 
			3 & ENET   & 0.398 \\  
			4 & ERFNet-DKS   & 0.408 \\ 
			5 & ERFNet-SAD   & 0.409 \\ 
			6 & ERFNet-IntRA-KD  &0.432 \\ 
			7 &\textbf{ Ours} & \textbf{0.512} \\
			\hline
		\end{tabular}%
		\label{comp}%
	\end{table}%
	
	We find intersection over union (IOU) between ground truth and predicted label image to validate our algorithm. We have also calculated precision and recall for better understanding of our results. We have beaten the state of art \cite{yuenanHou2019ENetSad}  on BDD100K lane segmentation dataset both in terms of accuracy(by 58\%) and mean IOU(by 87\%). Please see Table  \ref{tab:BDD100K Comparison} for results on BDD100K.

	\begin{table}[htbp]
		\small
		\centering
		\caption{Comparative results on BDD100K dataset}
		\begin{tabular}{|c|c|c|c|}
			\hline
			S No. &  $Method$ & $Accuracy$ & $IOU$  \\
			\hline\hline
			1  & ResNet-18 \cite{khe2016Resnet} & 30.66 & 11.07  \\
			2  & ResNet-34 \cite{khe2016Resnet}  & 30.92 & 12.24 \\
			3  & ResNet-101\cite{khe2016Resnet} & 34.45 & 15.02 \\
			4  & ENet \cite{adamPaszke2016Enet} & 34.12 & 14.64 \\
			5  & SCNN \cite{xingang2017SCNN} & 35.79 & 15.84 \\
			6  & ENet-SAD \cite{yuenanHou2019ENetSad}  & 36.56 &16.02 \\
			7  & \textbf{Ours}  & \textbf{57.2} & \textbf{30.0} \\
			\hline
		\end{tabular}%
		\label{tab:BDD100K Comparison}%
	\end{table}%

\begin{figure*}[htbp]
		\begin{center}
			\subfigure[]{\includegraphics[width=0.8\linewidth]{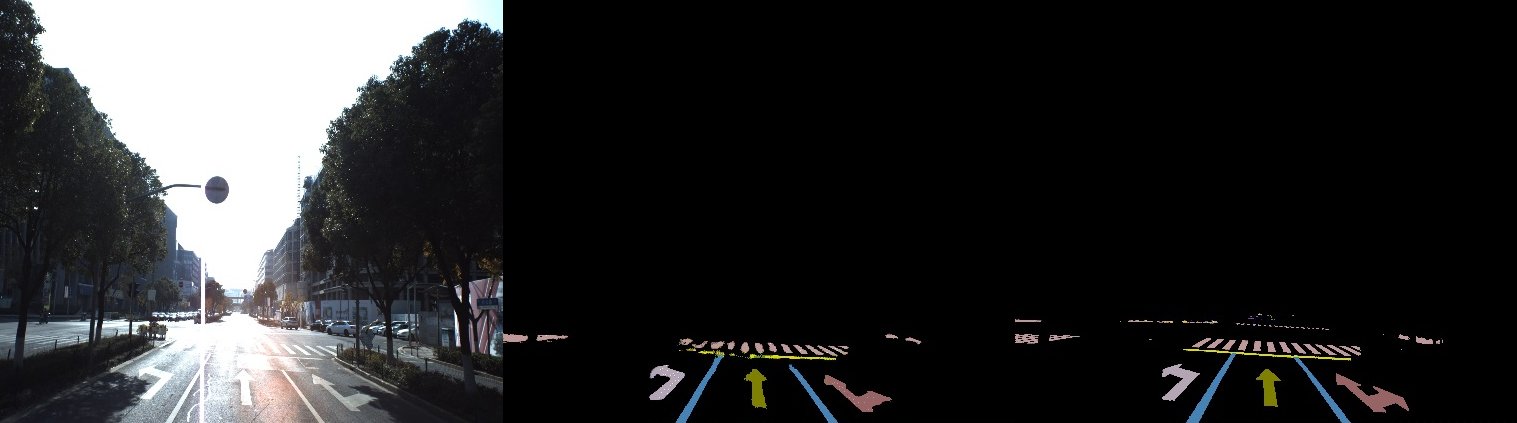}}
			\subfigure[]{\includegraphics[width=0.8\linewidth]{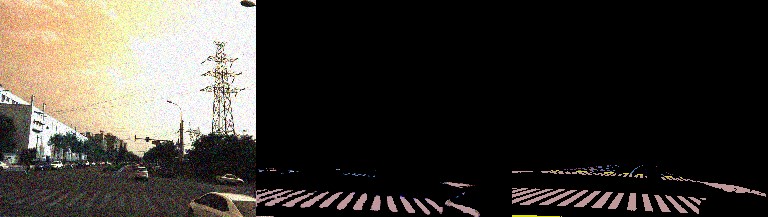}}
			\subfigure[]{\includegraphics[width=0.8\linewidth]{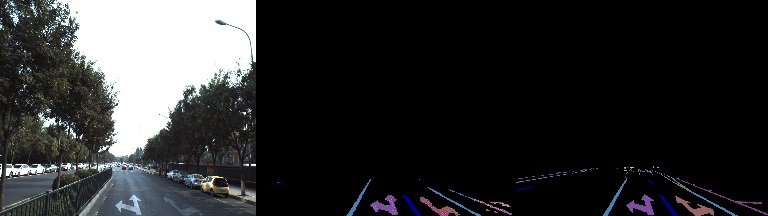}}
			\caption{{Examples of generated lanes and symbols under adverse conditions on ApolloScape. Left-Context Image, Center-Generated image, Right-Ground Truth}}
			\label{fig:sample_apollo_results_noisy}
		\end{center}
	\end{figure*}

\begin{figure*}[htbp]
	\begin{center}
		\subfigure{\includegraphics[width=0.8\linewidth]{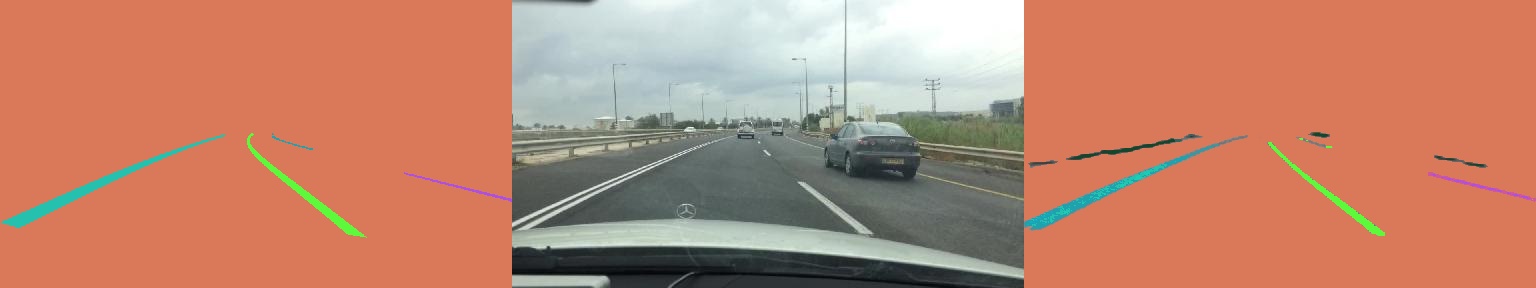}}
		\subfigure{\includegraphics[width=0.8\linewidth]{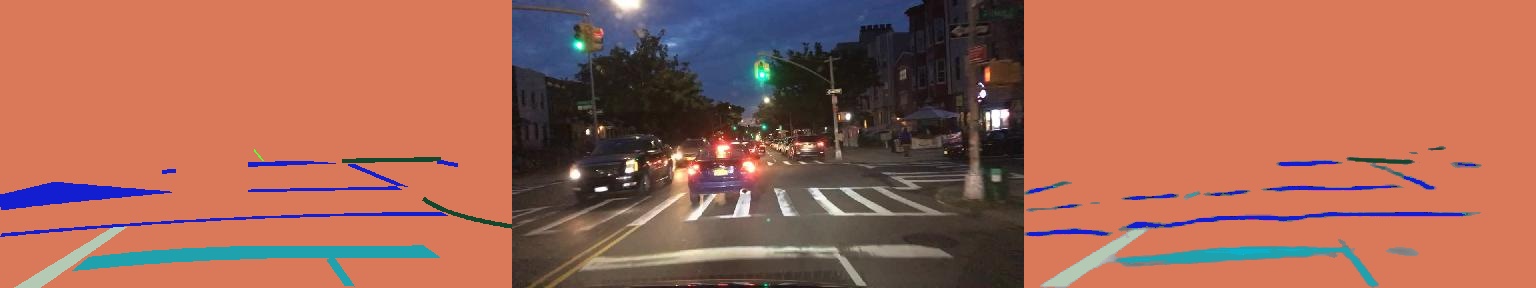}}
		\caption{{Examples of generated lanes and symbols on BDD100K. Left-Groudtruth, Center-Context Image, Right-Generated Image. 1st row, generated image shows that our model is able to generate curb even when its not in ground truth.}}
		\label{fig:sample bdd results}
	\end{center}
    \end{figure*}

We tested our algorithm on ApolloScape to ascertain if the model is capabale of handling variability on different types of road and symbols. As presented in Table \ref{tab:ApolloResults} we achieved 58\% IOU and 84\% precision on ApolloScape data. The results on both BDD100K and ApolloScape datasets validate our assumption that generative models perform better than most state of art discriminative models wherein probability distribution should be modelled. We can observe from Figure \ref{fig:sample_apollo_results_noisy} and Figure \ref{fig:sample bdd results} that our model is able to generate good results on various illumination and road contexts.

	\begin{table}[htbp]
		\small
		\centering
		\caption{Results on  ApolloScape data }
		\begin{tabular}{|c|c|c|c|c|}
			\hline
			S No. &  $Class$ & $IOU$ & $Precision$ & $ Recall $\\
			\hline\hline
			1  & Dividing lane & 0.81 & 0.93  & 0.86\\
			2  & Guiding lane  & 0.45 & 0.74 & 0.54\\
			3  & Crossing  & 0.69 & 0.92 & 0.74\\
			4  & Stop lane  & 0.60 & 0.9 & 0.85\\
			5  & Turn symbols  & 0.49 & 0.76 & 0.59\\
			6  & No parking  & 0.44 & 0.83 & 0.48\\
			7  & \textbf{Mean}  & \textbf{0.58} &\textbf{0.84} & \textbf{0.68}\\
			\hline
		\end{tabular}%
		\label{tab:ApolloResults}%
	\end{table}%

We did another experiment for to validate our assumption, that network is capable of generating lanes in adverse conditions. We divided the 3000K original test images into, three sets of 1000 images each. In one of the set we randomly introduce the Gaussian noise with zero mean while another set is subjected to illumination changes randomly using gamma correction. In the final set we have selectively removed some of lanes and symbols in an image by replacing them with neighborhood pixels in order to simulate occlusions and faded symbols. Altered dataset is again tested for validation metrics and results have been reported in Table \ref{tab:Results noisy}
	
	\begin{table}[htbp]
		\small
		\centering
		\caption{Results on ApolloScape data with adverse conditions}
		\begin{tabular}{|c|c|c|c|c|}
			\hline
			C		S No. &  $Class$ & $IOU$ & $Precision$&  $Recall$ \\
			\hline\hline
			1  & Dividing lane & 0.54 & 0.87  & 0.57\\
			2  & Guiding lane  & 0.32 & 0.72 & 0.37\\
			3  & Crossing  & 0.42 & 0.88 & 0.48\\
			4  & Stop lane  & 0.39 & 0.85 & 0.43\\
			5  & Turn symbols  & 0.31 & 0.66 & 0.45 \\
			6  & No parking  & 0.29 & 0.75 & 0.37\\
			7  & \textbf{Mean}  & \textbf{0.38} & \textbf{0.79} & \textbf{0.45}\\
			\hline
		\end{tabular}%
		\label{tab:Results noisy}%
	\end{table}%
We can observe from the Table \ref{tab:Results noisy}, there is low drop in accuracy which reinforces that the network is able to generalize and is robust in adverse conditions and occlusions. In Figure \ref{fig:sample_apollo_results_noisy} we have presented the results generated for images subjected to adverse conditions. Figure \ref{fig:sample_apollo_results_noisy}(a) represents sunglare drowning out the lanes and symbols however we can see that expected results have been generated. Similarly Figure \ref{fig:sample_apollo_results_noisy}(b) represents image where due to illumination and noise crosswalk is poorly visible. Figure \ref{fig:sample_apollo_results_noisy}(c) represents a scenario where in camera image lanes are faded away but our model is able to still generate lanes.

	\subsection{Ablation study}
	The approach we have proposed explores the idea of learning an implicit distribution through the use of a conditioned discriminator. However we also trained our network without adversarial loss to check how much effective is adversarial loss in helping the generator learn the distribution. We saw diminished performance on the same training setup without adversarial loss.Table\ref{tab:ablation} shows that without adversarial loss the network simply fails to learn distribution. 
	
\begin{figure*}[htbp]
		\begin{center}
			\subfigure[]{\includegraphics[width=0.8\linewidth]{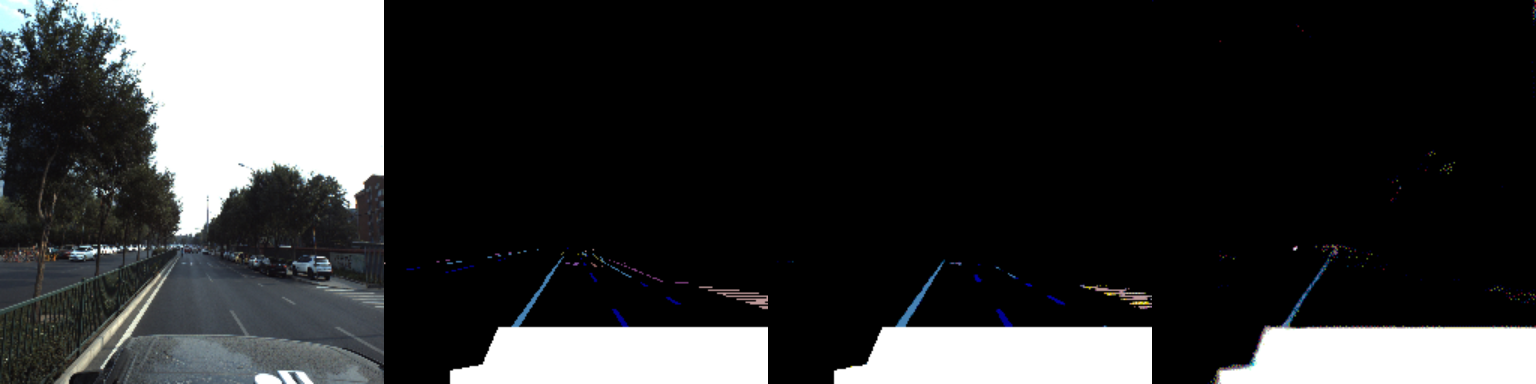}}
			\caption{{Comparison with and without adversarial loss. 1st column represents the context image. 2nd column  groundtruth. 3rd generated image with adversarial loss and 4th column represents generated image without adversarial loss}}
			\label{fig:compariosn_w_wo_adv}
		\end{center}
	\end{figure*}

	\begin{table}[htbp]
		\small
		\centering
		\caption{IOU of majority classes with and without adversarial loss}
		\begin{tabular}{|c|c|c|c|}
			\hline
			S No. &  $Class$ & $IOU$ $with$ $Adv$ & $IOU$ $w/o$ $adv$ \\
			\hline\hline
			1  & Dividing lane & 0.80  & 0.43\\
			2  & Guiding lane  & 0.45  & 0.14\\
			3  & Zebra         & 0.69  & 0.32\\
			4  & Stopping lane & 0.60  & 0.33\\
			3  & thru/turn     & 0.49  & 0.22\\
			\hline
		\end{tabular}%
		\label{tab:ablation}%
	\end{table}%
	A qualitative comparison between images generated using adversarial and without adversarial loss is shown in figure \ref{fig:compariosn_w_wo_adv}.

\subsection{Network complexity}
Proposed network has about 41 million trainable parameters with a model size of 200 MB and compares very well with other segmentation networks like PSPnet and SCNN which have about 100-150 million parameters. Model takes about 32 milliseconds for processing each frame of resolution $512 \times 512$ on a Intel Xeon platform with Nvidia TITAN RTX 2080 GPU onboard, Which is a real time performance. Model can be further optimized as the reported statistics are using Keras library which has considerable overhead.

\section{Conclusion and future work}
In the current paper we have proposed an algorithm for modeling the data distribution of road symbols and thereby generating them contextually. Proposed method has been trained and tested on Baidu's ApolloScape and BDD100K dataset. The network produces promising results along with low complexity when compared with the state of the art. In future we would like to extend our work to solve other problems related to perception for autonomous driving. Traffic lights and sign detection is one such problem where generative models can learn probability distribution and generalize better than discriminative models. Generatiev models can also be used to make robust perception algorithm that are resilient against adversarial attacks.

{\small
		\bibliographystyle{ieee_fullname}
		\bibliography{egbib}
}
	
\end{document}